\documentclass[letterpaper, 10 pt, conference]{ieeeconf}
\IEEEoverridecommandlockouts
\let\labelindent\relax
\usepackage{graphicx} \graphicspath{ {figures/} }
\usepackage{amsmath,amssymb,mathabx}
\usepackage{algorithmic}
\usepackage[ruled]{algorithm2e}
\usepackage{acronym}
\usepackage{enumitem}
\usepackage{booktabs}
\usepackage{hyperref}
\usepackage{balance}
\usepackage{xspace,setspace}
\usepackage[skip=3pt,font=small]{subcaption}
\usepackage[skip=3pt,font=small]{caption}
\usepackage[dvipsnames]{xcolor}
\usepackage[capitalise]{cleveref}
\usepackage{tabularx,colortbl,multirow,array,makecell}
\usepackage{overpic}
\usepackage{cite}
\usepackage{tikz}

\newcommand*\circled[1]{\tikz[baseline=(char.base)]{
            \node[shape=circle,draw,inner sep=0.6pt] (char) {#1};}}
            
\makeatletter
\DeclareRobustCommand\onedot{\futurelet\@let@token\@onedot}
\def\@onedot{\ifx\@let@token.\else.\null\fi\xspace}
\def\eg{\emph{e.g}\onedot} 
\def\ie{\emph{i.e}\onedot} 
 
\def\etc{\emph{etc}\onedot} 
\def\wrt{w.r.t\onedot} 
\def\etal{\emph{et al}\onedot}
\makeatother

\crefname{algocf}{alg.}{algs.}
\Crefname{algocf}{Algorithm}{Algorithms}

\makeatletter
\def\BState{\State\hskip-\ALG@thistlm}
\makeatother

\makeatletter
\renewcommand{\paragraph}{%
  \@startsection{paragraph}{4}%
  {\z@}{0ex \@plus 0ex \@minus 0ex}{-1em}%
  {\hskip\parindent\normalfont\normalsize\bfseries}%
}
\makeatother

\crefname{algocf}{alg.}{algs.}
\Crefname{algocf}{Algorithm}{Algorithms}

\definecolor{gblue}{HTML}{4285F4}
\definecolor{gred}{HTML}{DB4437}

\acrodef{dof}[DoF]{Degree of Freedom}
\acrodef{vkc}[VKC]{Virtual Kinematic Chain}
\acrodef{tamp}[TAMP]{Task and Motion Planning}
\acrodef{pddl}[PDDL]{Planning Domain Definition Language}

\title{\LARGE{}\bf{}\fontsize{14.5}{14.5}\selectfont{}Consolidating Kinematic Models to Promote Coordinated Mobile Manipulations}

\author{Ziyuan Jiao$^{1*}$\quad{}Zeyu Zhang$^{1*}$\quad{}Xin Jiang$^{3}$\quad{}David Han$^{2}$\quad{}Song-Chun Zhu$^{1}$\quad{}Yixin Zhu$^{1}$\quad{}Hangxin Liu$^{1}$
\thanks{$^{*}$ Ziyuan Jiao and Zeyu Zhang contributed equally to this work.}
\thanks{$^{1}$ UCLA Center for Vision, Cognition, Learning, and Autonomy (VCLA) at Statistics Department. Emails:
        {\tt\{zyjiao, zeyuzhang, yixin.zhu, hx.liu\}@ucla.edu},
        \tt{sczhu@stat.ucla.edu}.}%
\thanks{$^{2}$ Drexel University, Department of Electrical and Computer Engineering.
Email:
        \tt{dkh42@drexel.edu}.}%
\thanks{$^{3}$ UCLA ECE Department.
Email:
        \tt{jiangxjames@ucla.edu}.}%
\thanks{The work reported herein was supported by ONR N00014-19-1-2153, ONR MURI N00014-16-1-2007, and DARPA XAI N66001-17-2-4029.}%
}

\begin{document}

\maketitle
\thispagestyle{empty}
\pagestyle{empty}

\begin{abstract}
We construct a \acf{vkc} that readily consolidates the kinematics of the mobile base, the arm, and the object to be manipulated in mobile manipulations. Accordingly, a mobile manipulation task is represented by altering the state of the constructed \ac{vkc}, which can be converted to a motion planning problem, formulated and solved by trajectory optimization. This new \ac{vkc} perspective of mobile manipulation allows a service robot to (i) produce well-coordinated motions, suitable for complex household environments, and (ii) perform intricate multi-step tasks while interacting with multiple objects without an explicit definition of intermediate goals. In simulated experiments, we validate these advantages by comparing the \ac{vkc}-based approach with baselines that solely optimize individual components. The results manifest that \ac{vkc}-based joint modeling and planning promote task success rates and produce more efficient trajectories.
\end{abstract}

\setstretch{0.97}

\section{Introduction}

Mobile manipulation is a core capability for a service robot to function properly in household environments and excel in various tasks. However, since service robots usually come with a bulky mobile base along with a large base footprint, they often struggle in household environments due to three unique challenges: (i) An indoor space is confined and clustered, constraining the robot's locomotion and posing additional challenges for mobile manipulation. (ii) The majority of tasks involve manipulating objects with diverse structures (\eg, articulated objects like doors and drawers), challenging to generate motion plans for mobile manipulators with a unified approach. (iii) The base and arm's movements have to be coordinated to ensure efficient and safe operations during mobile manipulation.

There exists research efforts focusing on each of the above challenges, \ie, motion planning in confined space~\cite{buchanan2019walking,cheong2020relocate,han2020towards}, control or learning frameworks for opening doors and drawers~\cite{jain2010pulling,karayiannidis2012open,karayiannidis2016adaptive,wang2020learning}, and whole-body planning for foot-arm coordination~\cite{shankar2016kinematics,bodily2017motion,chitta2010planning}. However, a unified approach that tackles all three challenges altogether in household settings is still mostly missing. Consequently, a service robot's mobile manipulation skills are far from ideal in terms of efficacy or fluency.

In stark contrast, humans possess fluid manipulation skills and interact with an environment efficiently. Cognitive psychologies and philosophers have proposed a theory of body schema~\cite{gallagher2006body}: Humans maintain a body's representation during their motions and interactions with the environment; this representation is malleable and can be extended to incorporate external objects. By treating the manipulated object as part of the extended limb, the theory of body schema provides a plausible account for why humans excel in complex manipulation tasks, from picking and placing an object, to opening doors and drawers, to tool-use~\cite{holmes2006beyond}. Although the idea of the body schema has been introduced to the robotics community to represent robot structures and guide robot's behaviors~\cite{hoffmann2010body}, it has left untouched whether the theory of body schema would promote a service robot's (mobile manipulation in particular) planning and execution skills in complex manipulation tasks. And if it does, what would be a proper representation at a computational level?

\begin{figure}[t!]
    \centering
    \includegraphics[width=\linewidth]{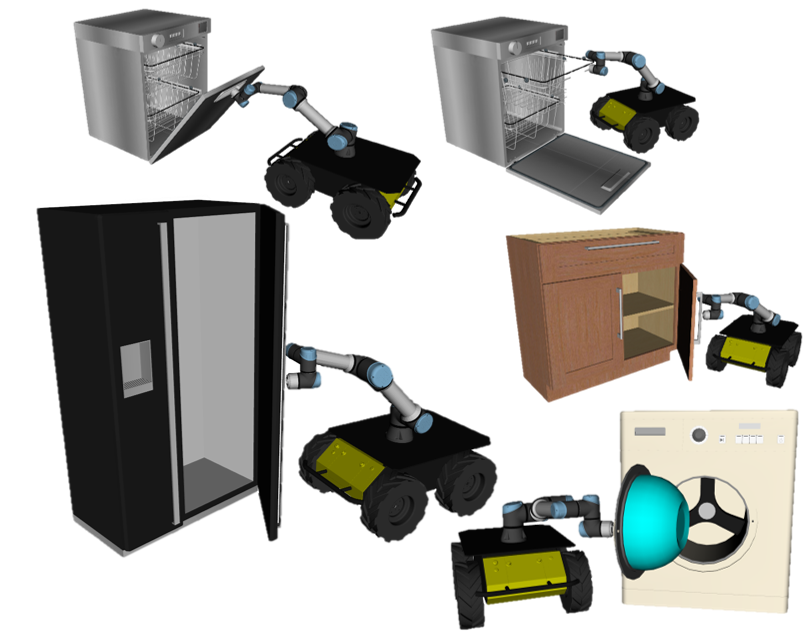}
    \caption{\textbf{Diverse interactions a service robot needs to perform in a household environment.} By abstracting the objects' kinematic structures and forming a \ac{vkc}, a service robot can plan and act more efficiently with improved foot-arm coordination.}
    \label{fig:motivation}
\end{figure}

To answer this question, we propose to abstract the object being manipulated---rigid objects and constrained mechanisms (doors, drawers, \etc)---by its kinematic structure. We integrate the kinematics of the robot and the manipulated object by constructing a single kinematic chain using the idea of \acf{vkc}~\cite{likar2014virtual}. Specifically, the kinematic structure of the manipulated object is augmented to the manipulator by connecting the end-effector to an attachable location of the object. A virtual transformation between the mobile base and the virtual base link is also augmented by incorporating its navigation information.

From this new \ac{vkc}-based perspective, a mobile manipulation task in a household environment is represented by altering the state or the structure of the \ac{vkc}, which leads to a motion planning problem on \ac{vkc}, formulated and solved by trajectory optimization. This new perspective enables a service robot to plan and act efficiently by allowing it to directly incorporate external objects and plan the motion as a whole to achieve better foot-arm coordination; see examples in \cref{fig:motivation}. In simulations, we validate the proposed \ac{vkc} perspective in various mobile manipulation tasks. Our experiments show that the consolidated kinematic models are particularly suitable for service robots by alleviating intermediate goal definitions for motion planners; they promote coordinated motions among base, arm, and object.

\subsection{Related Work}\label{sec:related_work}

The idea of \textbf{\acf{vkc}} could be traced back to 1997 by Pratt \etal~\cite{pratt1997virtual} for bipedal robot locomotion~\cite{pratt2001virtual}. This idea was later adopted to chain serial manipulators to form one kinematic chain~\cite{likar2014virtual} and to dual-arm manipulation tasks; for instance, connecting parallel structures via rigid-body objects~\cite{wang2015cooperative}, modeling whole-body control of mobile manipulators~\cite{wang2016whole}. Recently, \ac{vkc} is also adopted for wheeled-legged robot control~\cite{laurenzi2019augmented}. In this paper, we further push the idea of \ac{vkc} to a mobile manipulator and demonstrate its advantages in modeling and planning complex manipulation tasks in household environments.

\textbf{Motion planning} is among the largest and most fundamental fields in robotics. In essence, methods can be roughly categorized into three major doctrines: search-based (\eg, A$^\star$~\cite{hart1968formal}, D$^\star$~\cite{stentz1997optimal}), sampling-based (\eg, RRT~\cite{lavalle2000rapidly} and its variants~\cite{kuffner2000rrt,karaman2010incremental}), and trajectory optimization (\eg, CHOMP~\cite{ratliff2009chomp}, TrajOpt~\cite{schulman2014motion}). We formulate the motion planning problem on \ac{vkc} following the conventions in TrajOpt, as it incorporates kinematic constraints better than sampling-based methods while avoiding searching in large spaces.

Efficiently performing \textbf{mobile manipulation} tasks are challenging. Notable efforts have recently been dedicated to algorithms or system implementations, focusing on interactive manipulation tasks. For instance, equilibrium point control~\cite{jain2010pulling} and impedance control structure~\cite{stuede2019door} are introduced to open doors and drawers. To improve efficiency, Gochev \etal used a heuristic-based method to reduce the search space~\cite{gochev2012planning}. Taking advantage of solving the inverse kinematics, Burget \etal proposed a whole-body motion planning approach for humanoid's constrained motion~\cite{burget2013whole}, and Bodily \etal proposed an algorithm for jointly optimizing a robot's base position and joint motions~\cite{bodily2017motion}. More recently, Toussaint \etal proposed a multi-bounded tree search algorithm to solve multi-step manipulation tasks involving tool-use~\cite{toussaint2018differentiable}. Despite their promising results, prior arts primarily focus on a specific problem setup (\eg, opening door and drawer, using tools). In comparison, the proposed approach rethinks mobile manipulation from a more general viewpoint using \ac{vkc}s and tackles a broader range of tasks.

\subsection{Overview}

The remainder of this paper is organized as follows. \cref{sec:definition} outlines notations and formally presents the problem definition. \cref{sec:modeling} details the \ac{vkc} modeling. In a series of mobile manipulation tasks, we demonstrate the efficacy of \ac{vkc}s with a high success rate in \cref{sec:exp}. We conclude the paper with discussions in \cref{sec:conclusion}.

\setstretch{1}

\begin{figure}[t!]
    \centering
    \includegraphics[width=0.95\linewidth]{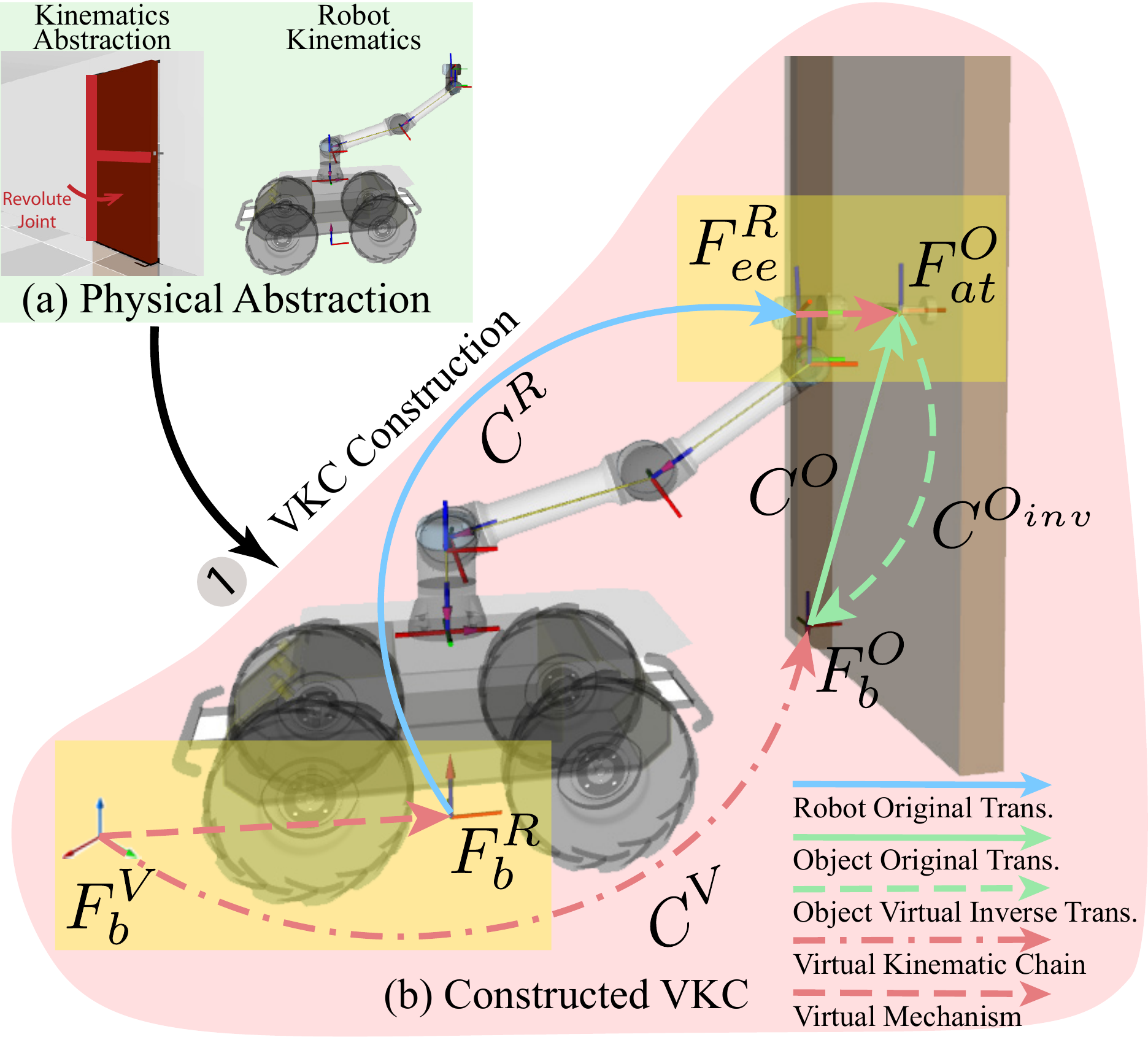}
    \caption{\textbf{Overview of the mobile manipulation planning schematics using the proposed \ac{vkc}-based approach.} (a) After abstracting out the underlying kinematics of the manipulated object and the mobile manipulator, (b) a \ac{vkc} is constructed. The yellow boxes denote where the virtual connections are established: (i) One between $\mathcal{F}^{V}_{b}$ and $\mathcal{F}^{R}_{b}$, the \emph{virtual} base frame in the world coordinate and the robot's actual base frame, to reflect the navigational information, and (ii) another between $\mathcal{F}^{R}_{ee}$ and $\mathcal{F}^{O}_{at}$, the robot's end-effector frame and the attachable frame of the object, to transfer effects of the manipulator to the manipulated object.}
    \label{fig:framework}
\end{figure}

\section{Notations and Problem Definition}\label{sec:definition}

This section introduces the notations throughout the paper and the problem setup describing a mobile manipulation task. 

The physical properties and kinematics of links and joints are defined following the Unified Robot Description Format (URDF) in Robot Operating System (ROS) and organized in a tree representation $\mathcal{T}$. \cref{tab:notations} lists all the related notations:
\begin{table}[ht!]
    \centering
    \caption{Notations used for constructing \ac{vkc}s.}
    \label{tab:notations}
    \resizebox{\linewidth}{!}{%
        \begin{tabular}{@{\hskip0pt}c@{\hskip3pt}c@{\hskip3pt}l@{\hskip0pt}}
            \hline
            \textbf{Group} & \textbf{Notation} & \textbf{\quad{}\quad{}\quad{}\quad{}\quad{}\quad{}Description}\\ \hline \hline
            \multirow{5}{*}{\rotatebox[origin=c]{90}{Robot}}
                &$\mathcal{T}^R$ & A tree represents the robot kinematic model\\
                &$\mathcal{F}^R_{b}$ & Robot base link's frame; the root of $\mathcal{C}^R$\\
                &$\mathcal{F}^R_{ee}$& Robot end-effector link's frame \\
                &$\mathcal{C}^R$ & $\subset \mathcal{T}^R$, a kinematic chain from $\mathcal{F}^R_{b}$ to $\mathcal{F}^R_{ee}$\\  
                &$\mathcal{F}^R_i$& Frame of link $i$ in the kinematic chain $\mathcal{C}^R$\\\hline
            \multirow{5}{*}{\rotatebox[origin=c]{90}{Object}}
                &$\mathcal{T}^O$ & A tree represents the object kinematic model\\ 
                &$\mathcal{F}^O_{b}$ & Object base link's frame; the root of $\mathcal{T}^O$\\
                &$\mathcal{F}^O_{at}$& Object attachable link's frame\\
                &$\mathcal{C}^O$ & $\subset\mathcal{T}^O$, a kinematic chain from $\mathcal{F}^O_{b}$ to $\mathcal{F}^O_{at}$\\ 
                &$\mathcal{F}^O_i$& Frame of link $i$ in the kinematic chain $\mathcal{C}^O$\\\hline
            \multirow{4}{*}{\rotatebox[origin=c]{90}{Others}}     &$\mathcal{C}^V_n$& A serial \ac{vkc} with $n$ \ac{dof} \\
            &$\mathbf{q}$& $\in\mathbb{R}^n$, the state of \ac{vkc} in joint space\\
            &$\mathbf{g}$& $\in\mathbb{R}^k$\ ($k\leq n$), the joint goal state\\
            &${}^{a}_{b}{T}$& 
            A homogeneous transformation from $\mathcal{F}_a$ to $\mathcal{F}_b$\\
            &${}^{w}_{i}{T}_{g}$& The goal pose of $\mathcal{F}_{i}$ in the world frame \\
            \hline
        \end{tabular}%
    }%
\end{table}

Below, we further summarize the above notations:
\begin{itemize}[leftmargin=*,noitemsep,nolistsep,topsep=0pt]
    \item The group \emph{Robot} refers to notations related to the mobile manipulator, which consists of three components: mobile base, manipulator, and end-effector.
    \item The group \emph{Object} refers to notations related to the manipulated objects, which could be as simple as a \textit{rigid} link or be an \textit{articulated} object with two or more links connected by either a prismatic, revolute, or fixed joint.
    We introduce a \emph{virtual} joint defined as an \emph{attachment}, a local transformation $^{at}_{ee}{T}$ from the object's attachable frame $\mathcal{F}^O_{at}$ (\ie, the link a mobile manipulator can grasp on) to the robot's end-effector frame $\mathcal{F}^R_{ee}$. 
    \item The group \emph{Others} refers to constructed \ac{vkc}, its state space, and other related notations in a manipulation task.
\end{itemize}

Constructing a \ac{vkc} $\mathcal{C}^V$ requires the inputs of robot kinematic tree $\mathcal{T}^R$, object kinematic tree $\mathcal{T}^O$, and transformation from an object attachable frame to the robot end-effector frame ${}^{at}_{ee}{T}$. The chain's forward kinematics (FK), inverse kinematics (IK), and Jacobians can be effectively solved by existing kinematic solvers (\eg, KDL~\cite{smits2011kdl}).

Assuming a rigid connection between the end-effector and the attachable link during manipulation, performing a mobile manipulation task can be regarded as reaching desired \ac{vkc} poses. As a result, we treat a mobile manipulation task as a motion planning problem on the \ac{vkc} and solve it by trajectory optimization. Formally, it is equivalent to finding a collision-free path $\mathbf{q}_{1:T}$ from the initial pose $\mathbf{q}_{init}$ to goals $\mathbf{g}$ in joint space and/or goal poses ${}^{w}_{i}{T}_{g}$ in Euclidean space.

The objective function of the trajectory optimization can be formally expressed as:
\begin{equation}
    \underset{\mathbf{q}_{1:T}}{\min} \sum_{t=1}^{T-1} || W_{vel}^{1/2} \;\; \delta{\mathbf{q}}_{t} ||_2^2
    \; + \sum_{t=2}^{T-1} || W_{acc}^{1/2} \;\; \delta\dot{\mathbf{q}}_{t} ||_2^2,
    \label{eqn:objective}
\end{equation}
wherein we penalize the overall weighted squared traveled distance of every joint with the finite forward difference $\delta{\mathbf{q}}_{t} \approx \mathbf{q}_{t+1} - \mathbf{q}_{t}$ and overall smoothness of the trajectory with the second-order finite central difference $\delta\dot{\mathbf{q}}_{t} \approx \mathbf{q}_{t-1} - 2\mathbf{q}_t + \mathbf{q}_{t+1}$. $W_{vel}$ and $W_{acc}$ are diagonal weight matrices for each joint, respectively. $\mathbf{q}_{1:T}$ represents the trajectory sequence $\{q_1, q_2, \ldots, q_T\}$, where $\mathbf{q}_{t}$ denotes the \ac{vkc} state at the $t^\mathrm{th}$ time step.

\section{\texorpdfstring{\ac{vkc}}{} Modeling}\label{sec:modeling}

The proposed \ac{vkc} modeling constructs a serial kinematic chain by (i) incorporating both robot and object kinematics via a virtual joint and (ii) augmenting a virtual base to the robot base; see \cref{fig:framework}b for a graphic illustration.

\subsection{\texorpdfstring{\ac{vkc}}{} Construction}

Below we formally describe the 4-step procedure of constructing the \ac{vkc}, $\mathcal{C}^V$, by consolidating the robot and the object kinematics models.

\paragraph{Original Structure}

The kinematic models of the mobile manipulator $\mathcal{T}^R$ and the manipulated object $\mathcal{T}^O$ are assumed given by the perception module or by the simulator.

\paragraph{Kinematic Inversion} 

Let us take the task of opening a door as an example. In conventional kinematic notation, the door is the child link, and the door frame is its parent link in the original $\mathcal{T}^O$. To construct a \ac{vkc}, this parent-child relationship needs to be inverted before it can be attached to the robot's end-effector, \ie, the door becomes the parent link that ``transforms'' the door frame. Of note, such an inversion also requires updating the joint connecting the two links, since a joint (\ie, revolute/prismatic) typically constrains the child link's motion \wrt the \emph{child link's} frame.

\paragraph{\ac{vkc} Construction}

After inverting the original $\mathcal{T}^O$, a virtual joint between $\mathcal{F}^{O_\mathrm{inv}}_{b}$ and $\mathcal{F}^{R}_{ee}$ is inserted, whose transformation is denoted as $^{ee}_{at}{T}$. In our application, the transformation of the virtual joint is updated by the actual grasping pose right before the \ac{vkc} construction to minimize kinematic discrepancies introduced by the execution error. Next, the motion planner will be invoked to plan following motions for the actual \ac{vkc}. The joint type could also be determined by the grasping type between the gripper and the object (\eg, revolute joint for grasping a cylindrical handle, fixed joint for grasping a rigid ball) to alleviate the inaccuracies during the execution.

\paragraph{Virtual Base Frame}

A virtual base frame $\mathcal{F}^{V}_{b}$ is further added and connected to the mobile base through two perpendicular prismatic joints and a revolute joint, enabling the mobile base's omnidirectional motions on the ground plane.

After the above procedure, the constructed \ac{vkc} remains in serial and forms an equality constraint to \cref{eqn:objective}: 
\begin{equation}
    h_{\text{chain}}(\mathbf{q}_t) = 0, \; \forall t = 1, 2, \ldots, T \label{eqn:chain_cnt}
\end{equation}
It specifies the kinematics of the \ac{vkc}, which includes its forward kinematics and other physical constraints of the manipulated object; \eg, the manipulated object is fixed to the ground, which leads to a closed chain: ${}^{w}_{b}{T}^{O}_{1:T}-{}^{w}_{b}{T}^{O}=0$. Failing to account for this constraint may damage the manipulated object or the mobile manipulator.

\subsection{Goals}

The goal of the mobile manipulation can be formulated as an inequality constraint, in addition to the equality constraint introduced by the \ac{vkc} construction in \cref{eqn:chain_cnt}:
\begin{equation}
    || f_{\text{task}}(\mathbf{q}_T) - \mathbf{g} ||^2_2 \leq \xi_{\text{goal}}, \label{eqn:goal_cnt}
\end{equation}
which bounds the squared $l2$ norm between the final state in the goal space $f_{\text{task}}(\mathbf{q}_T)$ and the goal state $\mathbf{g}$ with a tolerance $\xi_{\text{goal}}$. The function $f_{\text{task}}: \mathbb{R}^n\rightarrow\mathbb{R}^k$ is a task-dependent function that maps the joint space of a \ac{vkc} to the goal space that differs from task to task.

Again, let us take the example of opening a door. In the first phase when the robot is reaching the door handle, $f_{\text{task}}(\cdot)$ maps the joint space of a \ac{vkc} to the robot's end-effector pose. In this case, the goal $\mathbf{g}$ is the robot's end-effector pose ${}^{w}_{ee}{T}_{g}$, and \cref{eqn:goal_cnt} can be rewritten in a simplified form $|| f_{\text{fk}}(\mathbf{q}_T) - {}^{w}_{i}{T}_{g} ||^2_2 \leq \xi_{\text{goal}}$. In the second phase when the robot is opening the door, $f_{\text{task}}(\cdot)$ maps \ac{vkc}'s joint space to the joint of the door's revolute axis. Hence, $\mathbf{g}$ is merely the angle $\theta$ of the revolute joint, and the trajectory of the other joints in the \ac{vkc} are implicitly generated by the optimization process, together with obstacle avoidance and trajectory smoothing. Of note, \cref{eqn:chain_cnt,eqn:goal_cnt} are not the only forms of constraints that a \ac{vkc}-based approach can incorporate; in fact, it is straightforward to add additional task constraints to the same optimization problem in \cref{eqn:objective}, depending on various task-specific requirements.

\subsection{Additional Constraints}

During the trajectory optimization, we further impose several safety constraints. Without loss of generality, we assume an omnidirectional base and purely kinematic constraints in this paper. However, extra constraints, such as nonholonomic constraints for non-omnidirectional mobile bases or dynamic constraints for arms, could be formulated into the optimization problem by incorporating additional time, first-order, or second-order terms~\cite{rosmann2017kinodynamic}.
\begin{align}
    \mathbf{q}^{\min} \leq \mathbf{q}_t \leq \mathbf{q}^{\max}, &\, \;\;\; \forall t = 1, 2, \ldots, T \label{eqn:joint_limit} \\
    ||\delta{\mathbf{q}}_{t}||_\infty \leq \xi_{\text{vel}}, ||\delta\dot{\mathbf{q}}_{t}||_\infty \leq \xi_{\text{acc}}, &\, \;\;\; \forall t = 2, 3, \ldots, T-1 \label{eqn:acc} \\
    \sum_{i= 1}^{N_{\text{link}}} \sum_{j=1}^{N_{\text{obj}}} |\text{dist}_{\text{safe}} - &f_{\text{dist}}(L_i, O_j)|^{+} \leq \xi_{\text{dist}} \label{eqn:collision_1}, \\
    \sum_{i= 1}^{N_{\text{link}}} \sum_{j=1}^{N_{\text{link}}} |\text{dist}_{\text{safe}} - &f_{\text{dist}}(L_i, L_j)|^{+} \leq \xi_{\text{dist}} \label{eqn:collision_2}.
\end{align}
\cref{eqn:joint_limit} is an inequality constraint that defines joint limits, in which $\mathbf{q}^{\min}$ and $\mathbf{q}^{\max}$ specify the lower and upper bound of every joint, respectively. \cref{eqn:acc} is an inequality constraint that bounds the joint velocity by $\xi_{\text{vel}}$ and the joint acceleration by $\xi_{\text{acc}}$ to obtain a feasible trajectory that can be executed without saturation. $||\cdot||_{\infty}$ denotes the infinity norm.

\cref{eqn:collision_1,eqn:collision_2} are inequality constraints that check link-object collisions and link-link collisions, respectively, where $N_{\text{link}}$ and $N_{\text{obj}}$ are the number of links and the number of objects, respectively. $\text{dist}_{\text{safe}}$ is a pre-define safety distance, and $f_{\text{dist}}(\cdot)$ is a function that calculates the signed distance~\cite{schulman2014motion} between $i$-th link $L_i$ and $j$-th object $O_j$; the function $|\cdot|^{+}$ is defined as $|x|^{+} = \max(x, 0)$. 

The inequality constraints introduced by \cref{eqn:collision_1,eqn:collision_2} make the preceding optimization problem highly non-convex and unsolvable by a generic convex solver. In this paper, we approximate it by a sequence of convex problems~\cite{schulman2014motion}, solved by a sequential convex optimization method.

\subsection{Advantages}

As formally derived in the above sections, solving mobile manipulation as trajectory optimization using the proposed \ac{vkc}-based approach introduces two advantages:
\begin{enumerate}[leftmargin=*,noitemsep,nolistsep,topsep=0pt]
    \item \textbf{Eliminating unnecessary intermediate goals}. Let us use the example of opening a door: Only one goal---the door's angle to be opened to---is required. The final poses of the mobile base and the manipulator are directly produced during the trajectory optimization process without manually specifying unnecessary intermediate goals. Hence, the \ac{vkc}-based approach provides versatility and simplicity for modeling mobile manipulation tasks.
    \item \textbf{Coordinating locomotion and manipulation}. Using \ac{vkc}s, the trajectory optimization jointly generates trajectories of the mobile base and the manipulator, producing coordinated locomotion and manipulation, which is oftentimes challenging for conventional methods.
\end{enumerate}

These two advantages are crucial for a service robot operating in a household environment. Below, we demonstrate these advantages in a series of mobile manipulation tasks.

\setstretch{0.96}

\section{Motion Planning on \texorpdfstring{\ac{vkc}}{}s}\label{sec:exp}

In this section, we start with simulation setup, followed by the evaluation of the trajectory-optimization-based motion planning on \ac{vkc} from three perspectives: (i) the necessary trajectory initialization required by the motion planner, (ii) the improvement on base-arm coordination using the \ac{vkc} approach compared with traditional setups that have to plan their motions separately, and (iii) the capability of operating in a household environment and performing various tasks.

\subsection{Platform and Simulation Setup}\label{sec:exp_setup}

The service robot platform we adopted for testing is a Universal Robot UR5e manipulator mounted on a Clearpath Husky A200 UGA; see \cref{fig:framework}b for a graphic illustration. The simulation environment is an arena with $10\text{m}\times10\text{m}$ in size, discretized into $100\times100$ grids. All the experiments are conducted using a desktop with an Intel i7-9700K CPU, running with ROS-Industrial Tesseract~\cite{armstrong_2018}.

\subsection{Trajectory Initialization}

Although a gradient-descent-based algorithm can effectively solve the trajectory optimization problem on \ac{vkc}s, it may also be easily stuck at local minima near the given initial trajectory~\cite{ratliff2009chomp,schulman2014motion}. As a result, a proper trajectory initialization is favored to improve the optimization result. Two primary trajectory initialization methods~\cite{magyar2019timed} are:
\begin{enumerate}[leftmargin=*,noitemsep,nolistsep,topsep=0pt]
    \item \textbf{Stationary:} The trajectory $\mathbf{q}_{1:T}$ is initialized by way-points $\mathbf{q}_t$ that are the same as the initial pose $\mathbf{q}_{\text{init}}$.
    \item \textbf{Interpolated:} The trajectory $\mathbf{q}_{1:T}$ is initialized by way-points that are linearly interpolated between the initial pose $\mathbf{q}_{\text{init}}$ and the goal pose (not $\mathbf{g}$ or ${}^{w}_{ee}{T}_g$; see below).
\end{enumerate}

\begin{figure}[t!]
    \centering
    \begin{subfigure}[b]{0.333\linewidth}
        \includegraphics[width=\linewidth]{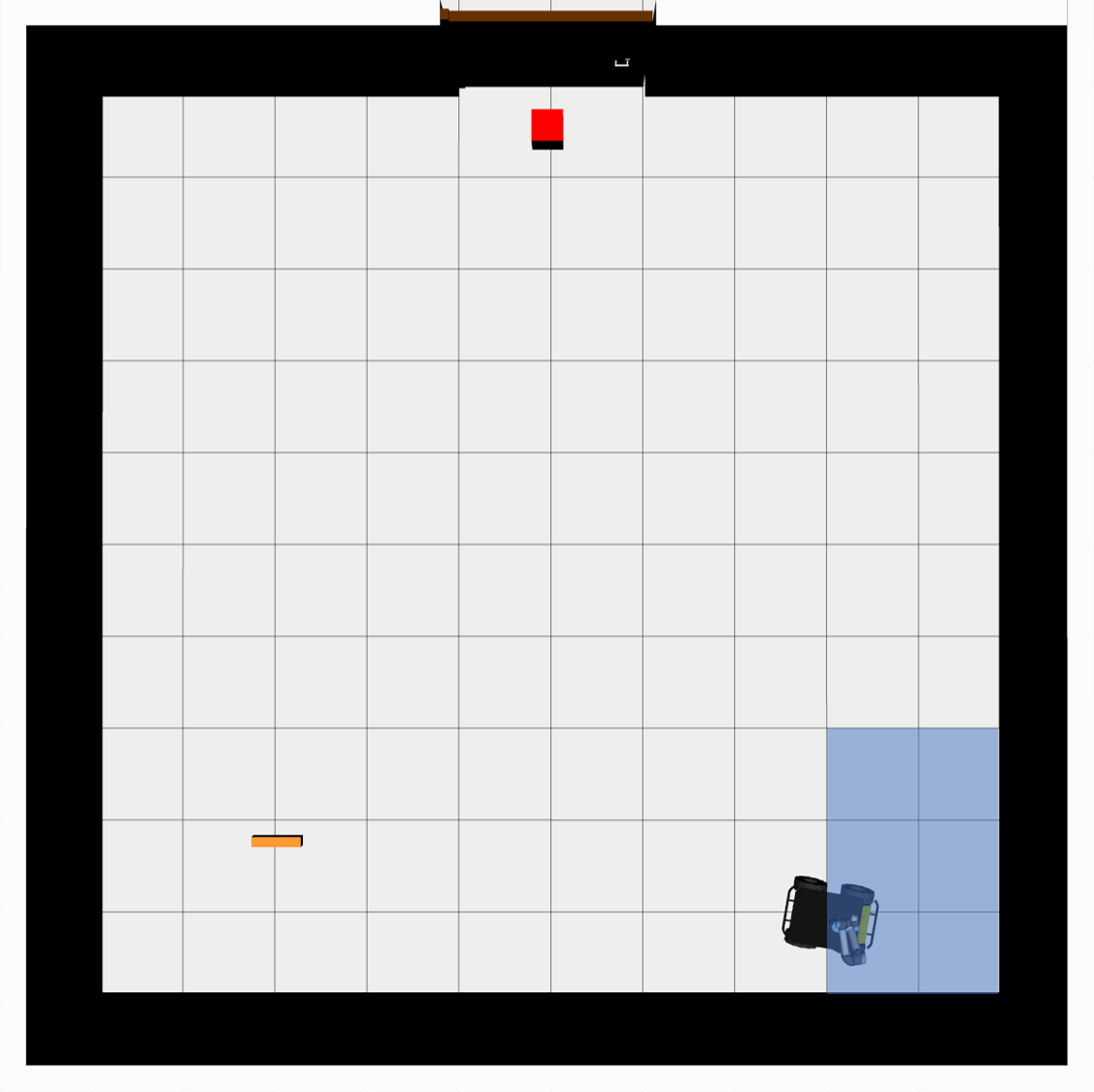}
        \caption{Scenario 1}
        \label{fig:exp_comparison_env1}
    \end{subfigure}%
    \begin{subfigure}[b]{0.333\linewidth}
        \includegraphics[width=\linewidth]{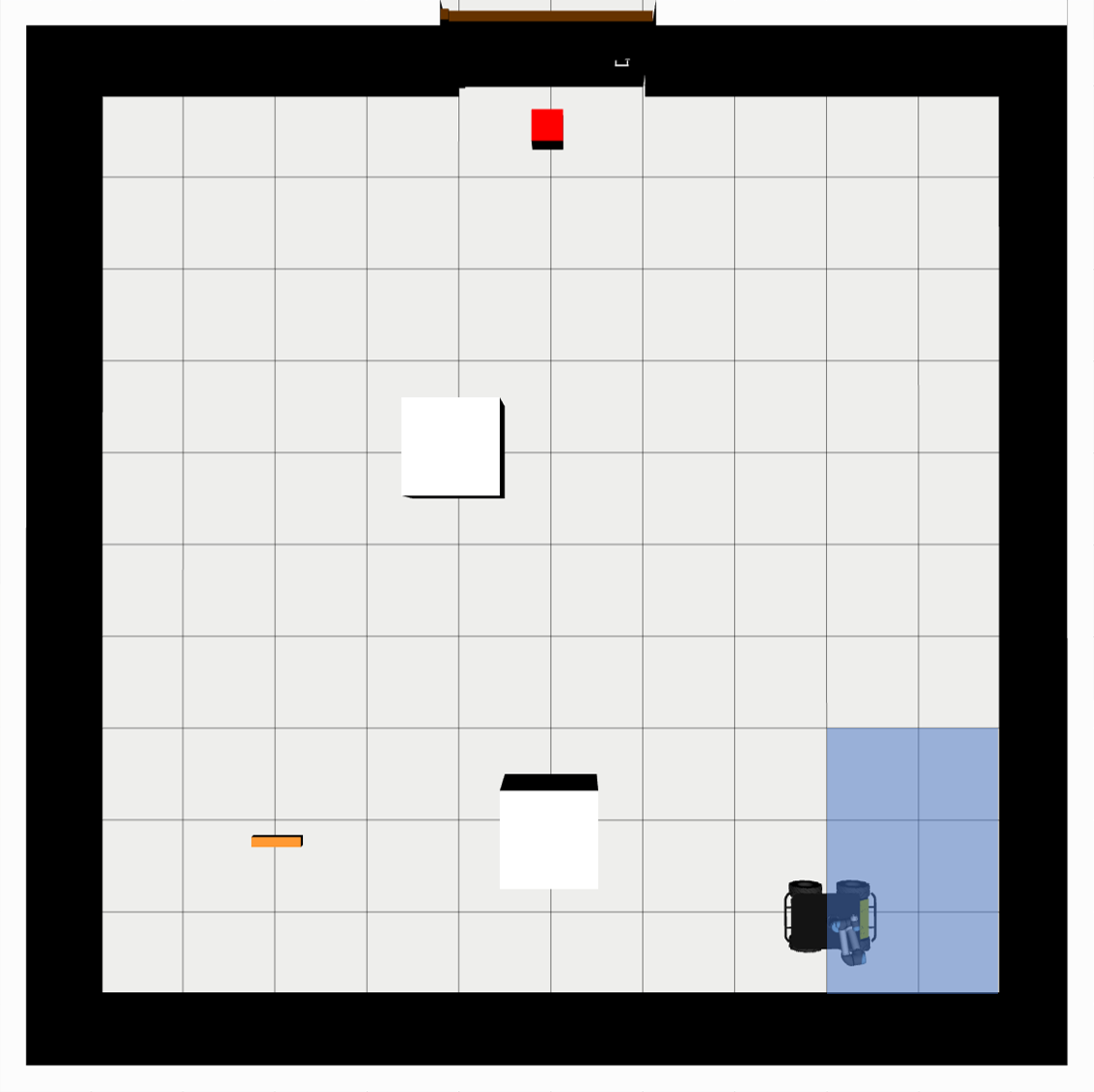}
        \caption{Scenario 2}
        \label{fig:exp_comparison_env2}
    \end{subfigure}%
    \begin{subfigure}[b]{0.333\linewidth}
        \includegraphics[width=\linewidth]{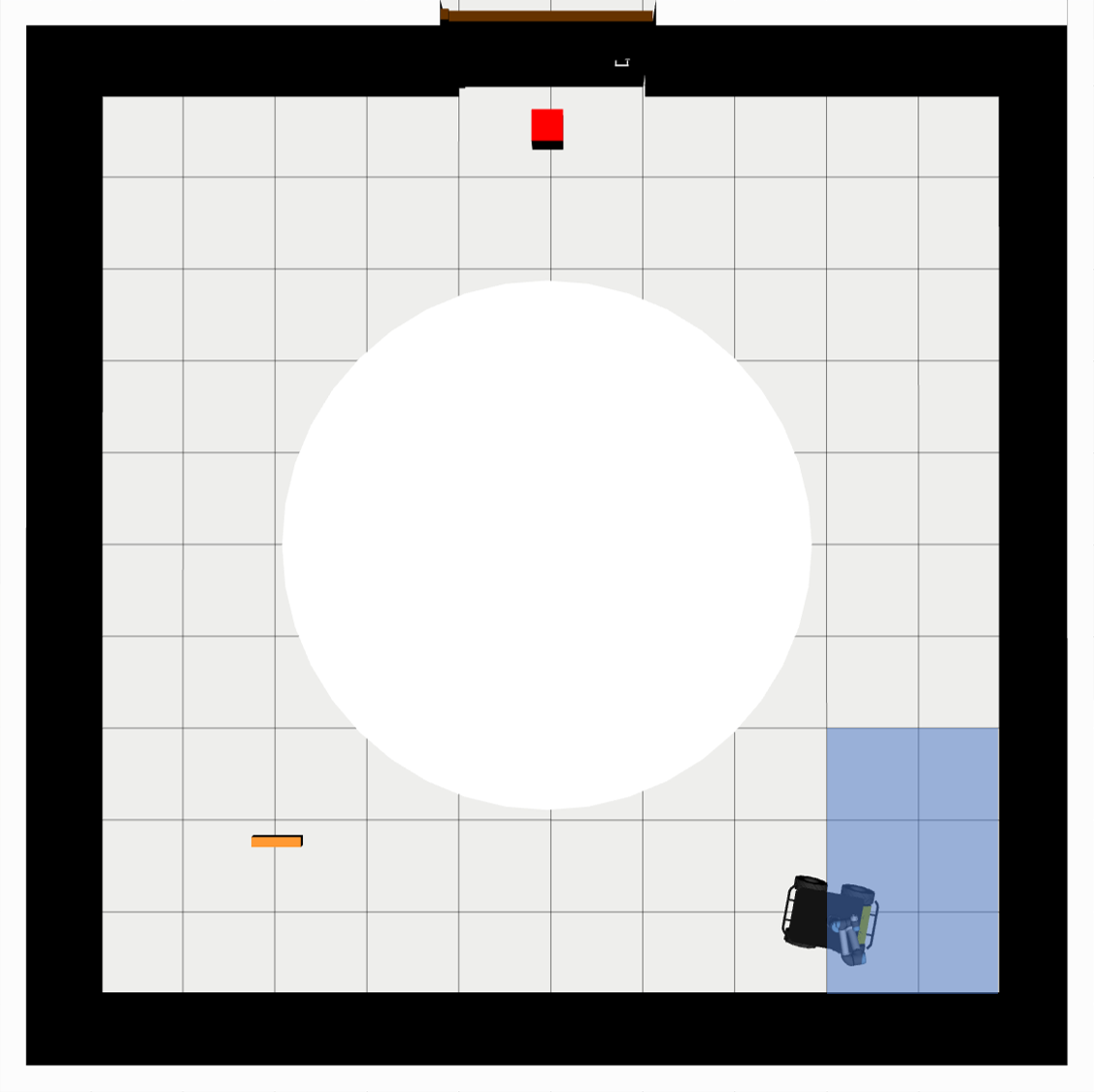}
        \caption{Scenario 3}
        \label{fig:exp_comparison_env3}
    \end{subfigure}%
    \\
    \begin{subfigure}[b]{\linewidth}
        \includegraphics[width=\linewidth]{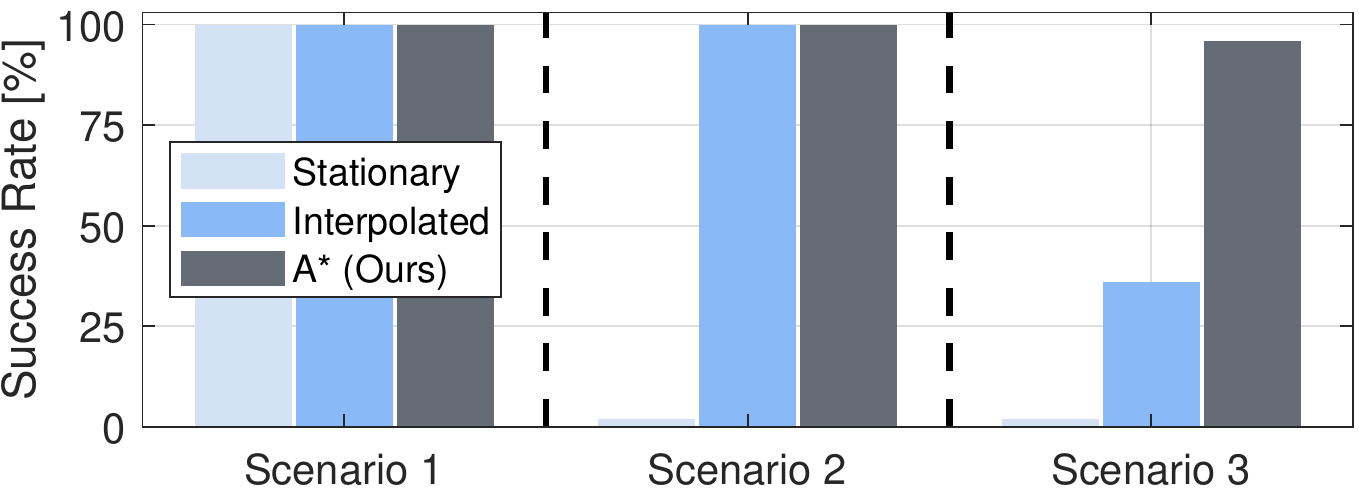}
        \caption{success rate}
        \label{fig:exp_comparison_success_rate}
    \end{subfigure}%
    \caption{\textbf{Comparisons of motion planning on \ac{vkc}s by different trajectory initialization methods.} (a)-(c) The experimental scenarios in increasing complexity. The robot's initial pose is uniformly sampled within the blue region; it is tasked to pick up the stick and use it to reach the red cube. (d) The proposed A$^\star$-based trajectory initialization has the highest success rates (almost always) in generating a feasible plan. In comparison, the \textit{Stationary} method fails to generate feasible plans in scenario 2 and 3. Similarly, the \textit{Interpolated} method struggles in scenarios 3 ($\tilde{}30\%$ success rate).
    } 
\end{figure}

In this paper, we further devise an A$^\star$-based trajectory initialization method, which adopts A$^\star$ to search for a feasible path given the initial and the goal pose of the mobile base. Next, we investigate how different trajectory initialization methods affect the planning results on \ac{vkc} in three 3D virtual scenarios; see \cref{fig:exp_comparison_env1,fig:exp_comparison_env2,fig:exp_comparison_env3}. The robot's task is to pick up the rigid stick and use it to reach a target indicated by the red cube. This task consists of three steps: navigate to the stick, manipulate the stick and pick it up, and navigate to and reach the target with the stick. The three scenarios designed for evaluation are in increasing complexity: no obstacle (\cref{fig:exp_comparison_env1}), two small obstacles (\cref{fig:exp_comparison_env2}), or a much larger one (\cref{fig:exp_comparison_env3}). Experimental results reported below are the average of 50 different initial poses, each with 10 times.

A successfully optimized trajectory is a converged result without violating any constraints (\eg, collisions). \cref{fig:exp_comparison_success_rate} compares success rates. When the environment is clean (Scenario 1), even the simplest \emph{Stationary} trajectory initialization method performs well. When there is additional complexity introduced by the obstacles (Scenario 2), the \emph{Stationary} method deteriorates, whereas the \emph{Interpolated} method still maintains a high success rate. When the navigable space is significantly reduced (Scenario 3), only the proposed \emph{A$^\star$}-based initialization method can consistently perform well to generate feasible plans. Taken together, experimental results indicate that combining the proposed \emph{A$^\star$}-based initialization with the optimization-based motion planner can well handle the challenging motions that require combining navigation and manipulation in cluttered space with obstacle avoidance.

\setstretch{0.95}

\subsection{Comparisons with Baselines}

\begin{figure}[t!]
    \centering
    \begin{subfigure}[b]{\linewidth}
        \caption*{Task 1: Push to open a door.}
        \begin{subfigure}[b]{0.25\linewidth}%
            \includegraphics[width=\linewidth]{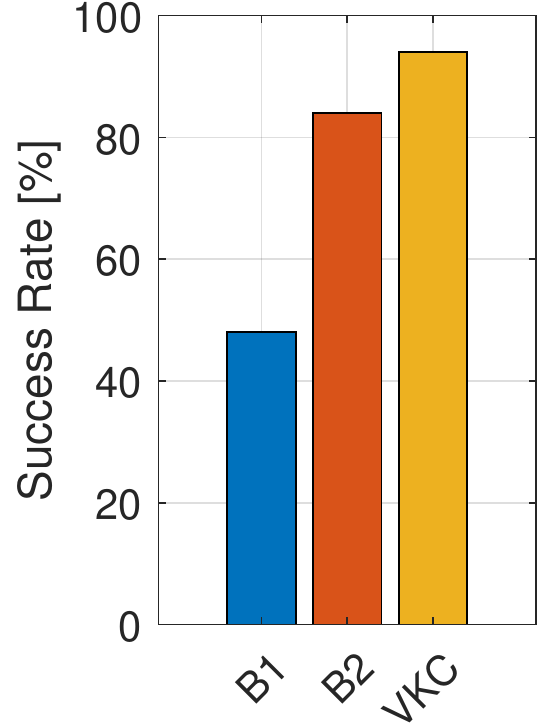}%
            \caption{succ. rate}%
        \end{subfigure}%
        \begin{subfigure}[b]{0.25\linewidth}%
            \includegraphics[width=\linewidth]{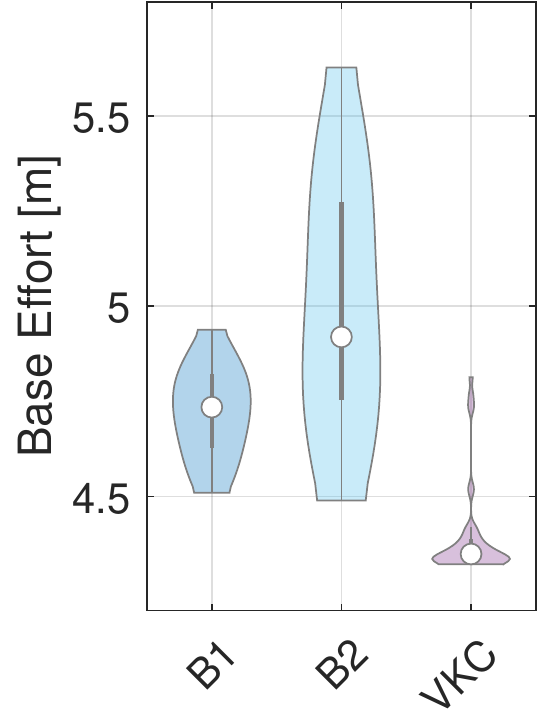}%
            \caption{base effort}%
        \end{subfigure}%
        \begin{subfigure}[b]{0.25\linewidth}%
            \includegraphics[width=\linewidth]{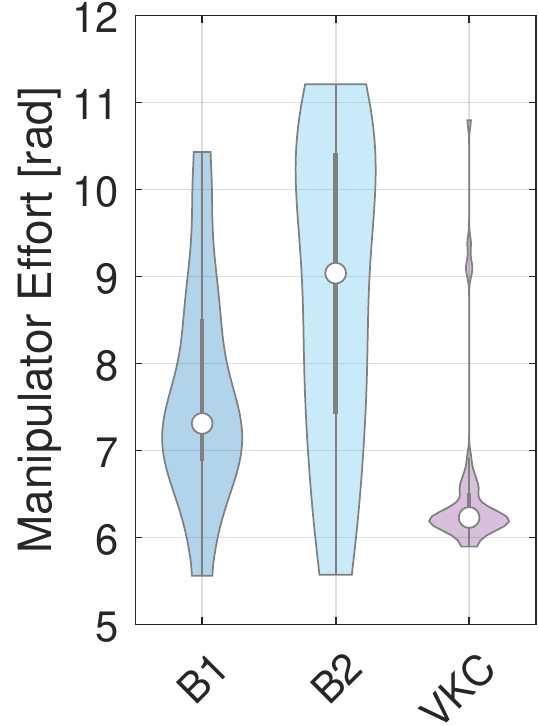}%
            \caption{mani. effort}%
        \end{subfigure}%
        \begin{subfigure}[b]{0.25\linewidth}%
            \includegraphics[width=\linewidth]{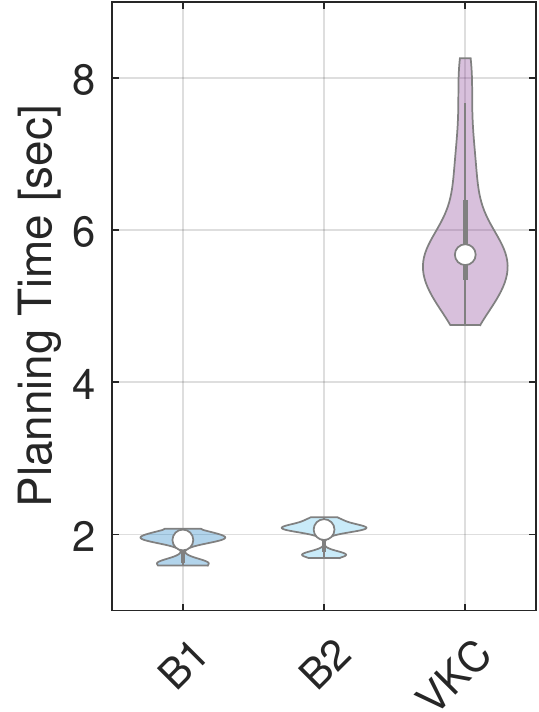}%
            \caption{time}%
        \end{subfigure}%
    \end{subfigure}%
    \\
    \begin{subfigure}[b]{0.25\linewidth}
        \includegraphics[height=\linewidth]{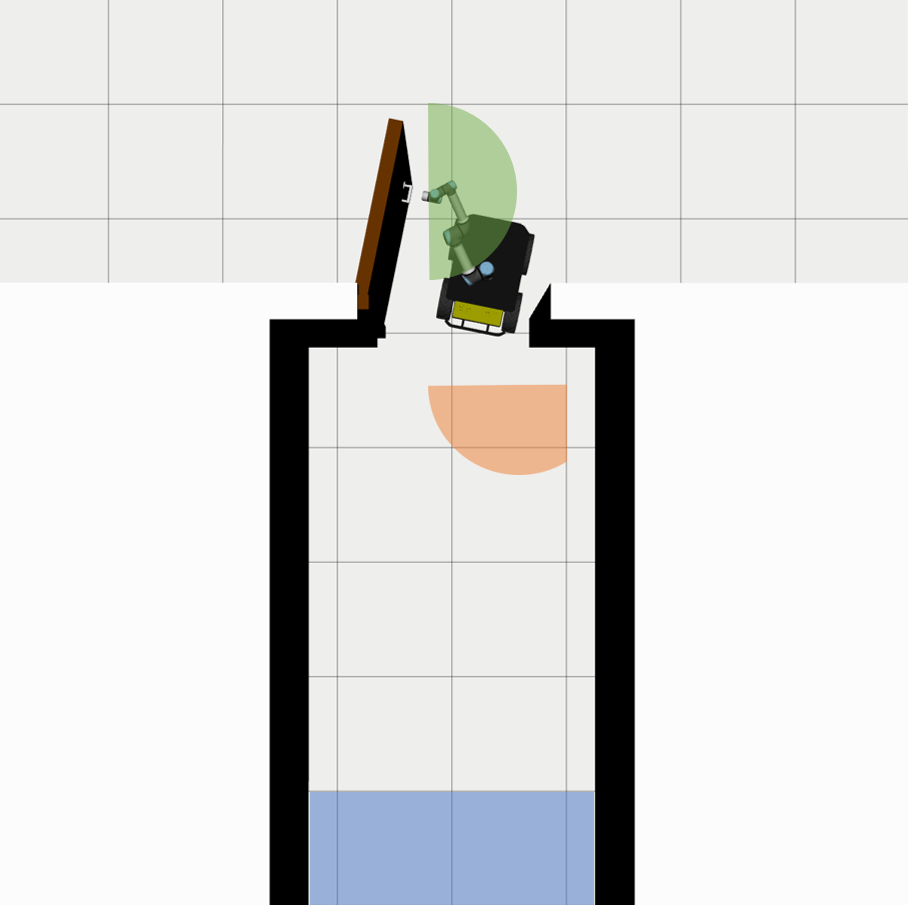}
        \caption{succ. case}%
        \label{fig:bsl_env1}
    \end{subfigure}%
    \begin{subfigure}[b]{0.25\linewidth}
        \includegraphics[height=\linewidth]{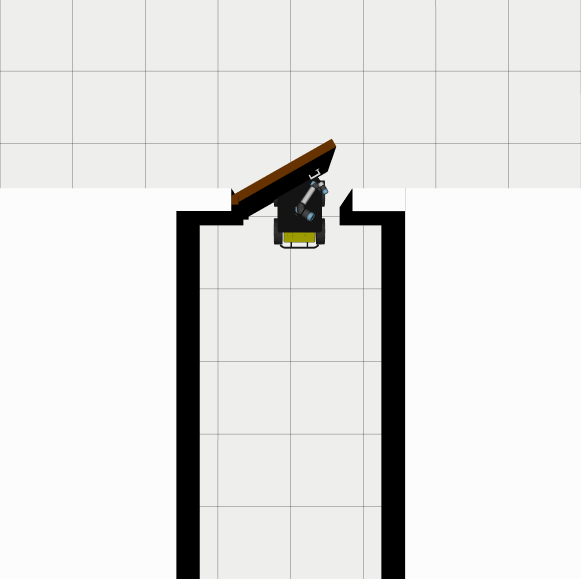}
        \caption{B1 fail. case}%
        \label{fig:bsl_env1_fail_B1}
    \end{subfigure}%
    \begin{subfigure}[b]{0.25\linewidth}
        \includegraphics[height=\linewidth]{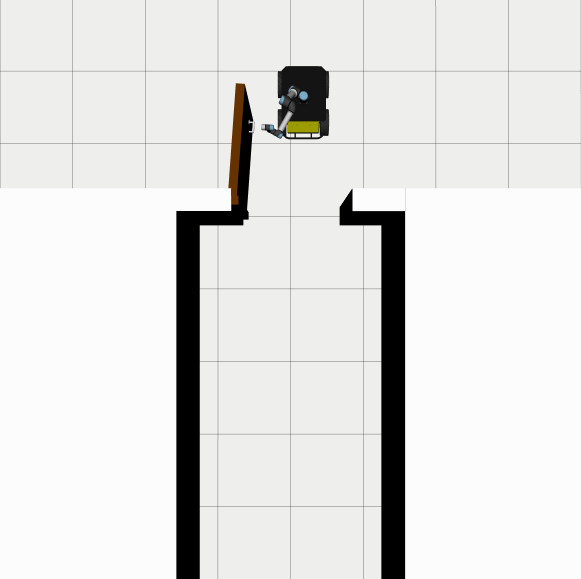}
        \caption{B2 fail. case}%
        \label{fig:bsl_env1_fail_B2}
    \end{subfigure}%
    \begin{subfigure}[b]{0.25\linewidth}
        \includegraphics[height=\linewidth]{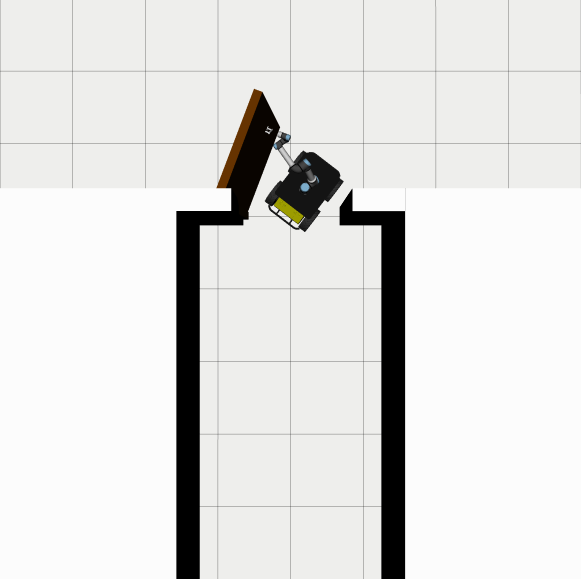}
        \caption{\fontsize{7}{7}\selectfont{}\ac{vkc} fail. case}%
        \label{fig:bsl_env1_fail_vkc}
    \end{subfigure}%
    \\\vspace{-9pt}\rule{\linewidth}{0.4pt}\vspace{-9pt}
    \\\phantom{}
    \begin{subfigure}[b]{\linewidth}
        \caption*{Task 2: Pull to open a drawer.}
        \begin{subfigure}[b]{0.25\linewidth}%
            \includegraphics[width=\linewidth]{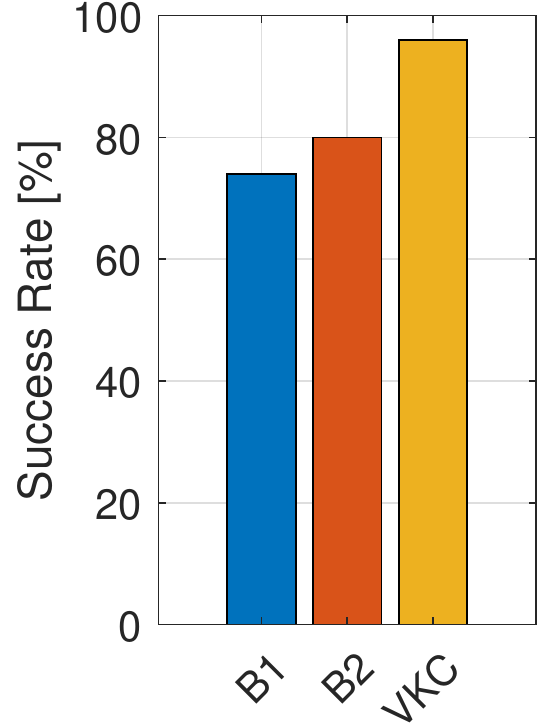}%
            \caption{succ. rate}%
        \end{subfigure}%
        \begin{subfigure}[b]{0.25\linewidth}%
            \includegraphics[width=\linewidth]{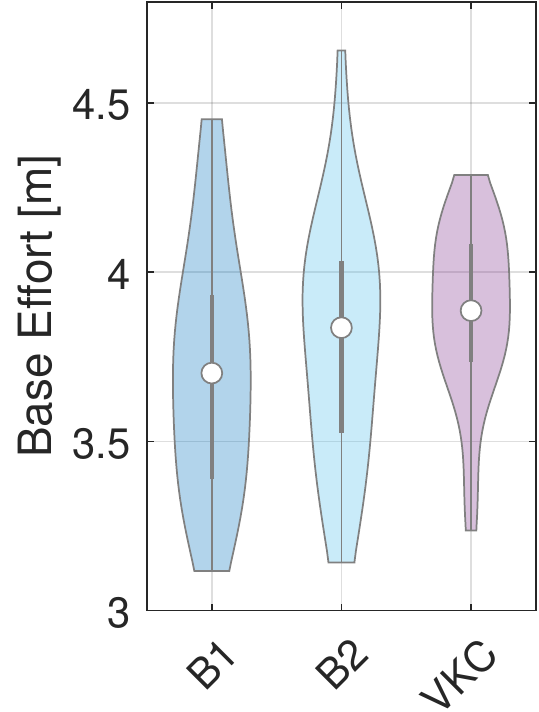}%
            \caption{base effort}%
        \end{subfigure}%
        \begin{subfigure}[b]{0.25\linewidth}%
            \includegraphics[width=\linewidth]{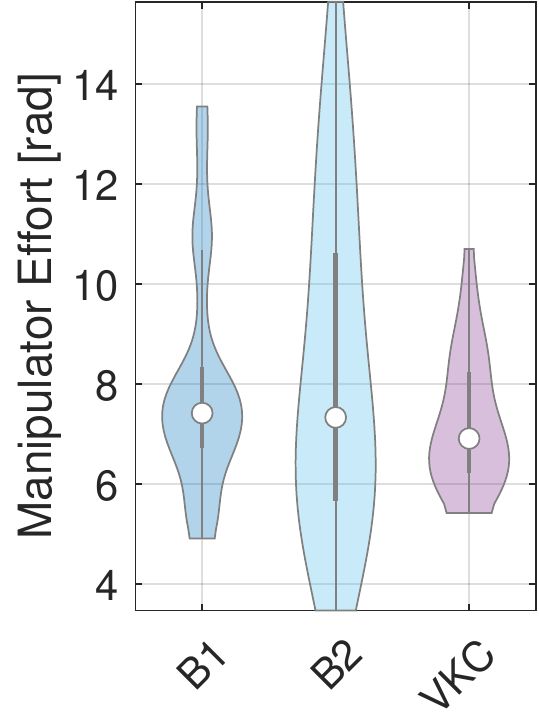}%
            \caption{mani. effort}%
        \end{subfigure}%
        \begin{subfigure}[b]{0.25\linewidth}%
            \includegraphics[width=\linewidth]{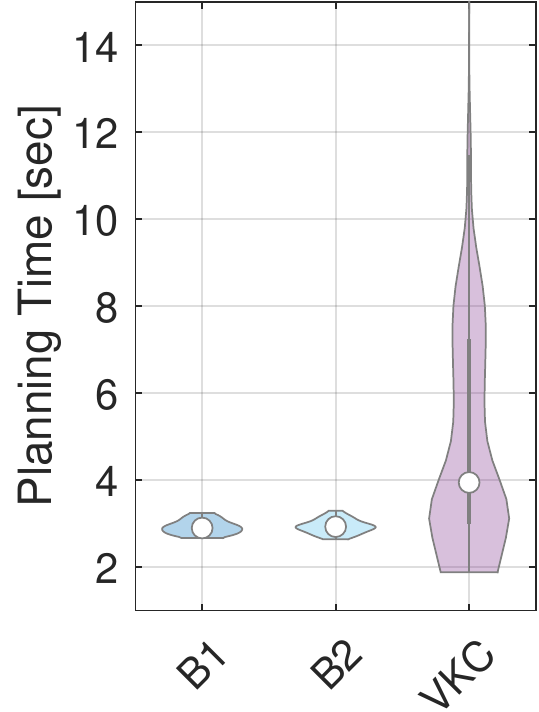}%
            \caption{time}%
        \end{subfigure}%
    \end{subfigure}%
    \\
    \begin{subfigure}[b]{0.25\linewidth}
        \includegraphics[width=\linewidth]{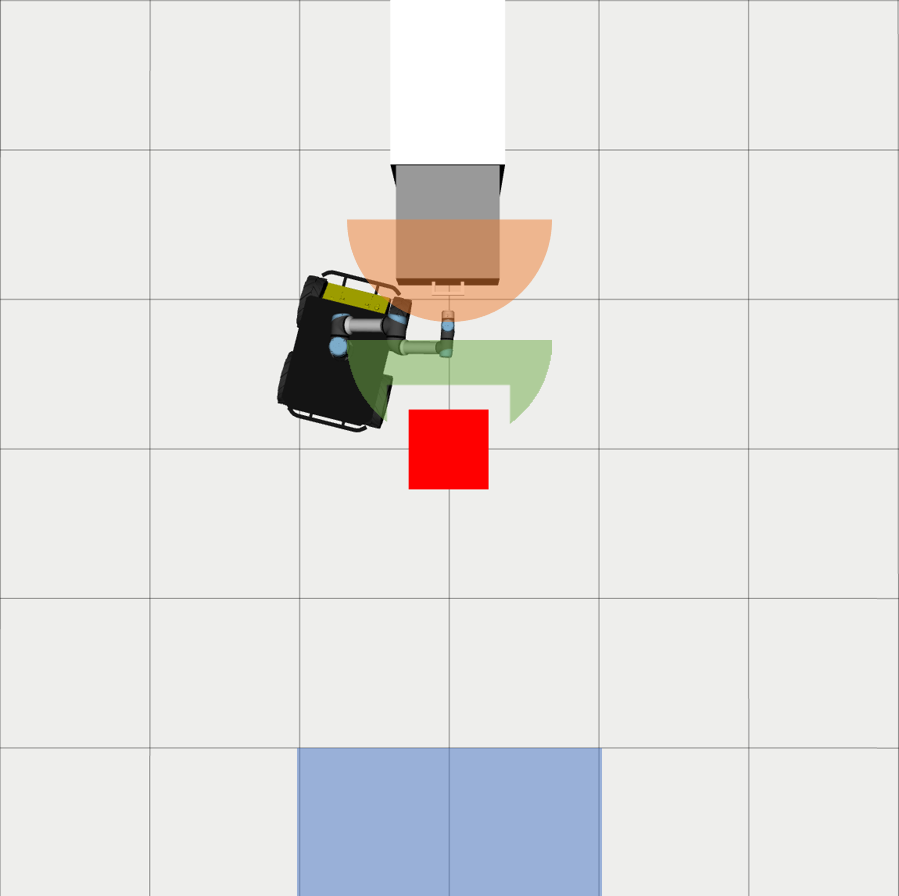}
        \caption{succ. case}%
        \label{fig:bsl_env2}
    \end{subfigure}%
    \begin{subfigure}[b]{0.25\linewidth}
        \includegraphics[width=\linewidth]{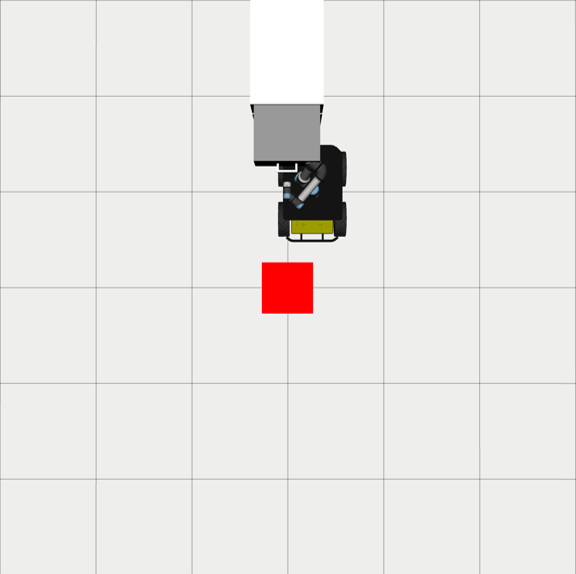}
        \caption{B1 fail. case}%
        \label{fig:bsl_env2_fail_B1}
    \end{subfigure}%
    \begin{subfigure}[b]{0.25\linewidth}
        \includegraphics[width=\linewidth]{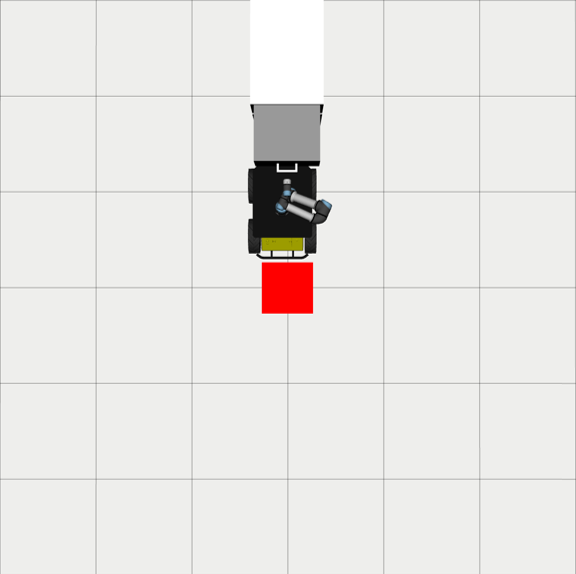}
        \caption{B2 fail. case}%
        \label{fig:bsl_env2_fail_B2}
    \end{subfigure}%
    \begin{subfigure}[b]{0.25\linewidth}
        \includegraphics[width=\linewidth]{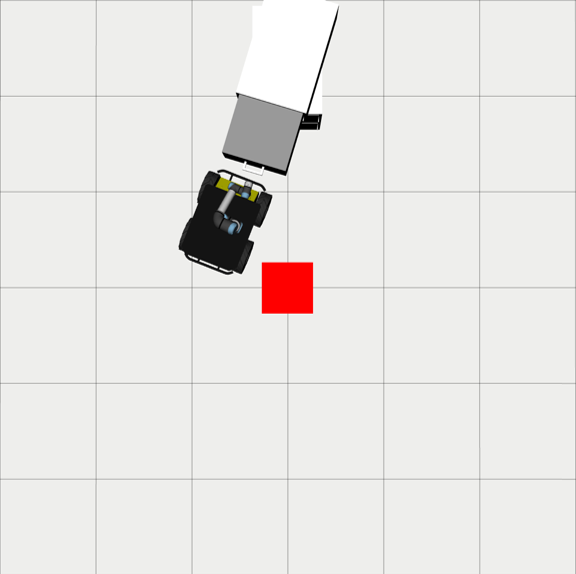}
        \caption{\fontsize{7}{7}\selectfont{}\ac{vkc} fail. case}%
        \label{fig:bsl_env2_fail_vkc}
    \end{subfigure}%
    \caption{\textbf{Quantitative comparisons between the \ac{vkc}-based approach and two baselines on two mobile manipulation tasks: push to open a door, and pull to open a drawer.} Modeling a mobile manipulation using \ac{vkc} leads to (a)(i) higher success rate and more effective trajectories in terms of smaller (b)(j) base and (c)(k) manipulator's efforts, but requires (d)(l) longer planning time. Of note, baselines require additional definition of intermediate goals, which are often empirically found, (e)(m) shown in the green and pink shaded areas. (f)(n) Typical failure cases of two tasks in B1 are due to collision. Typical failure cases of two tasks in (g)(o) B2 or (h)(p) \ac{vkc}-based approach are due to constraint violations.}
    \label{fig:baseline}
\end{figure}

We design two mobile manipulation tasks to validate the advantages of the proposed \ac{vkc} modeling: (i) rotating the doorknob and pushing to open a door (with two revolute joints), and (ii) pulling to open a drawer (with one prismatic joint); see \cref{fig:bsl_env1,fig:bsl_env2}. The robot starts from a uniformly sampled initial pose in the blue-shaded regions.

To benchmark the proposed \ac{vkc}-based approach, we design two baselines that have to \textit{individually} optimize the mobile base and mobile manipulator's trajectories. Baseline 1 (\textbf{B1}): Use the proposed A$^\star$-based trajectory initialization to first plan the mobile base's motion, and the arm pose is then found by solving the inverse kinematics from the door handle to the mobile base at each way-point. Baseline 2 (\textbf{B2}): On top of B1, the manipulated object and the manipulator's poses are further refined at each time step to avoid collisions between the mobile manipulator and the environment.

Of note, our \ac{vkc}-based approach only needs to specify one task goal---the desired door angle or the proper drawer length. In contrast, baselines require \textit{additionally} specifying the mobile base's pose when reaching the doorknob and after opening the door, as the base and the manipulator are planned and optimized individually. We compute these intermediate mobile base's poses by sampling from feasible regions that are empirically found; see the pink areas in \cref{fig:bsl_env1,fig:bsl_env2} for reaching and the green areas for final poses.

We evaluate the planning results by four criteria: (i) the percentage of task completion without violating the constraints as the \textit{success rate}, (ii) the total base traveling distance as the \textit{base's effort}, (iii) the sum of each joint's accumulated angular displacement during the entire task execution as the \textit{manipulator's effort}, and (iv) planning time.

We summarize results in \cref{fig:baseline}. Planning individually (\textbf{B1}) yields a 48\% and a 74\% success rate for opening door and drawer, respectively. The primary reason for failure is the collisions between the mobile manipulator and the door/drawer; see \cref{fig:bsl_env1_fail_B1,fig:bsl_env2_fail_B1}. Although introducing a collision check to refine motions (\textbf{B2}) improves the success rate, the proposed \ac{vkc}-based approach significantly outperforms the baselines. The failure cases of \textbf{B2} are mainly caused by violating velocity and acceleration constraints; see \cref{fig:bsl_env1_fail_B2,fig:bsl_env2_fail_B2}.

\begin{figure}[t!]
    \centering
    \includegraphics[width=\linewidth]{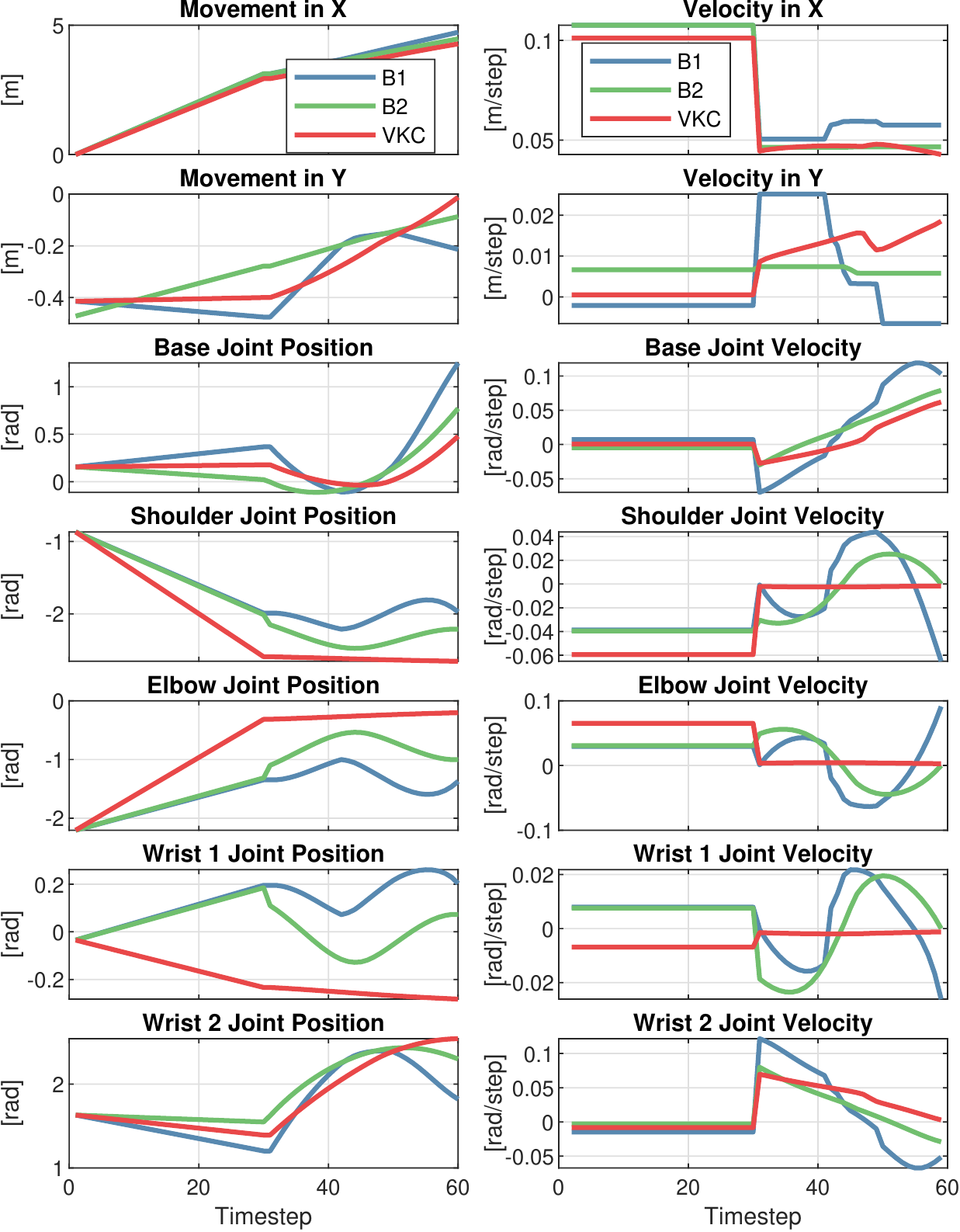}
    \caption{\textbf{The joint position and velocity profiles of the door-opening task in \cref{fig:bsl_env1}.} \ac{vkc}-based approach generates more efficient trajectory with less fluctuation compared to the baselines.}
    \label{fig:trajectory}
\end{figure}

Planning on \ac{vkc}s yields more efficient motions with less base and manipulator efforts; see an instance in \cref{fig:trajectory}.
Overall, the proposed \ac{vkc}-based approach produces smoother trajectories with less fluctuation in the speed profile.

\begin{figure*}[t!]
    \centering
    \includegraphics[width=\linewidth]{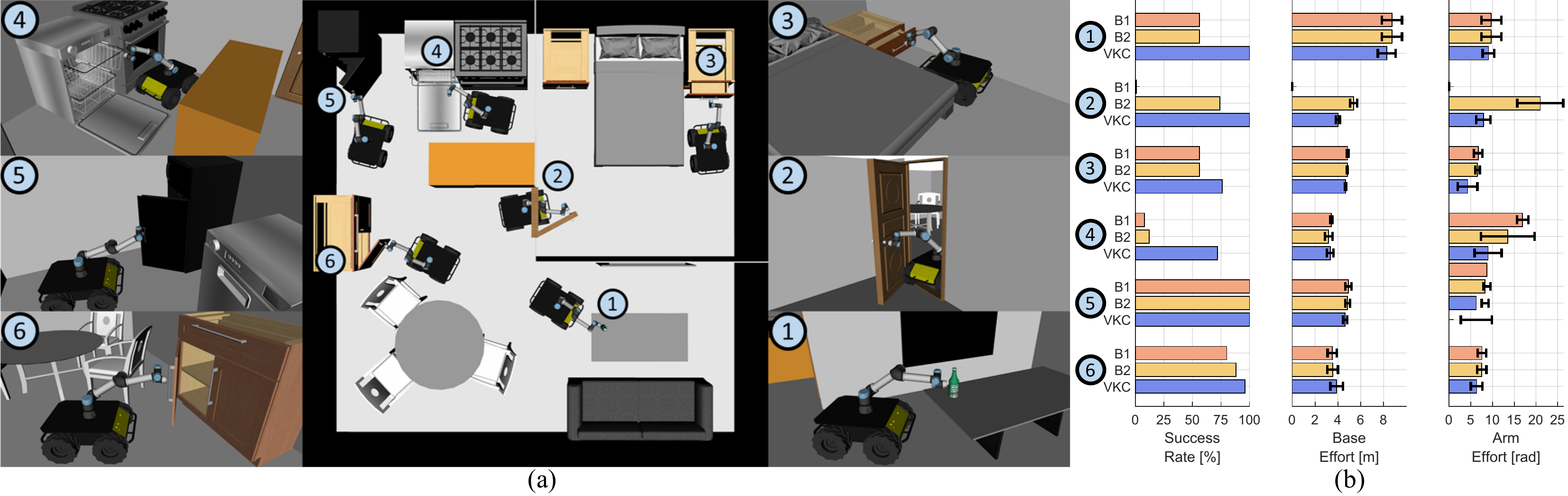}
    \caption{\textbf{The \ac{vkc}-based approach is advantageous in modeling and planning household mobile manipulation tasks for a service robot.} (a) Our simulated environments and the corresponding task performances. (b) The success rate, and base and arm efforts measured by their travel distance in 6 different mobile manipulation tasks. The black line segments denote the standard deviations. The proposed \ac{vkc}-based approach generally leads to a higher success rate with less base and arm efforts.}
    \label{fig:big_task}
\end{figure*}

Taken together, our experimental results demonstrate that by using the proposed \ac{vkc}-based approach, the planning of mobile manipulator avoids contrived definitions of intermediate poses and achieves a higher success rate with more efficient motions, at the cost of a slightly longer planning time due to solving the trajectory optimization problem with higher \ac{dof}s compared to those in the baselines.

\subsection{Complex Tasks}

We further test the \ac{vkc} modeling in a confined household environment with six daily manipulation tasks; see \cref{fig:big_task}a. 
\begin{enumerate}[leftmargin=*,noitemsep,nolistsep,topsep=0pt]
    \item Pick up an object randomly sampled on the coffee table (reachable by the robot) and place it on the dining table. \textbf{B2} performs the same as the \textbf{B1} since the object is fixed and does not collide with the robot.
    \item Push to open a narrow door. The robot needs to coordinate its foot and arm movements to open a narrow door (1.1 times wider than the robot) and pass through.
    \item Open a drawer in the corner. Although space is confined, the robot only needs a backward motion, which should be relatively straightforward.  
    \item Pull out a dishwasher rack, similar to the above task and also in a confined space. However, it is more challenging as the robot has to manipulate to its side.
    \item Pull to open a refrigerator.
    \item Pull to open a cabinet, similar to the above task. However, since the cabinet has a larger door size, the chair in the environment will block the robot's movement.
\end{enumerate}


In a successful trial, the robot should produce a collision-free trajectory without violating any kinematic or safety constraints. We compare the results in \cref{fig:big_task}b with two baselines (\textbf{B1} and \textbf{B2}). Overall, the proposed \ac{vkc}-based method demonstrates a higher success rate with less base and arm costs measured by the total distance traveled.

Specifically, in Task \circled{2}, \textbf{B1} can hardly find a path to go through the door because of the highly constrained free-space, which requires good foot-arm coordination; task \circled{4} reveals a similar result. In Task \circled{3}, the low drawer height increases the difficulty for the mobile base to reach a feasible region; hence, both baselines underperform in terms of their success rate. An interesting observation is in Task \circled{5} and \circled{6}: The baseline methods perform well in the pulling action, likely due to the empty space for the furniture doors to swing.

These results provide ample evidence of excellent foot-arm coordination produced by the proposed \ac{vkc}-based approach, well-suited for confined household environments.

\section{Conclusion and Discussion}\label{sec:conclusion}

We presented a modeling method that incorporates the kinematics of a robot's mobile base, arm, and the manipulated object in \ac{vkc}s. From this new perspective, a mobile manipulation task is regarded as a motion planning problem on \ac{vkc}s, solved by trajectory optimization. This approach alleviates the definition of intermediate goals and well coordinates base and arm movements, resulting in a higher success rate with more efficient trajectories in various mobile manipulation tasks. Our simulated experiments validate the advantages introduced by the proposed \ac{vkc}-based modeling approach, showing its potential in scaling up to complex operations in a household environment~\cite{jiao2021efficient}. 

we discuss two issues related to the presented work in greater depth. To deploy the \ac{vkc}-based method on a physical service robot, \textbf{objects' kinematic structures} have to be identified. Although some existing methods can achieve this goal by visual sensors~\cite{barragan2014interactive,paolillo2018interlinked,martin2019coupled,abbatematteo2019learning,han2021reconstructing}, results could be error-prone, and a force/torque control framework should be implemented to handle kinematic discrepancies or uncertainties/errors in object positions. Additionally, an effective and \textbf{stable grasp} is a prerequisite of constructing a \ac{vkc} connecting the robot and objects. While grasping remains an unsolved problem, more powerful machine learning-based methods are emerging~\cite{liang2019pointnetgpd,mousavian20196,ichnowski2020deep}, which could shed light on this direction.

\setstretch{0.88}

{
\tiny
\balance
\bibliographystyle{ieeetr}
\bibliography{IEEEfull}

\begin{thebibliography}{10}

\bibitem{buchanan2019walking}
R.~Buchanan, T.~Bandyopadhyay, M.~Bjelonic, L.~Wellhausen, M.~Hutter, and
  N.~Kottege, ``Walking posture adaptation for legged robot navigation in
  confined spaces,'' {\em Robotics and Automation Letters (RA-L)}, vol.~4,
  no.~2, pp.~2148--2155, 2019.

\bibitem{cheong2020relocate}
S.~H. Cheong, B.~Y. Cho, J.~Lee, C.~Kim, and C.~Nam, ``Where to relocate?:
  Object rearrangement inside cluttered and confined environments for robotic
  manipulation,'' in {\em Proceedings of International Conference on Robotics
  and Automation (ICRA)}, 2020.

\bibitem{han2020towards}
Z.~Han, J.~Allspaw, G.~LeMasurier, J.~Parrillo, D.~Giger, S.~R. Ahmadzadeh, and
  H.~A. Yanco, ``Towards mobile multi-task manipulation in a confined and
  integrated environment with irregular objects,'' in {\em Proceedings of
  International Conference on Robotics and Automation (ICRA)}, 2020.

\bibitem{jain2010pulling}
A.~Jain and C.~C. Kemp, ``Pulling open doors and drawers: Coordinating an
  omni-directional base and a compliant arm with equilibrium point control,''
  in {\em Proceedings of International Conference on Robotics and Automation
  (ICRA)}, 2010.

\bibitem{karayiannidis2012open}
Y.~Karayiannidis, C.~Smith, F.~E. Vina, P.~Ogren, and D.~Kragic, ``“open
  sesame!” adaptive force/velocity control for opening unknown doors,'' in
  {\em Proceedings of International Conference on Intelligent Robots and
  Systems (IROS)}, 2012.

\bibitem{karayiannidis2016adaptive}
Y.~Karayiannidis, C.~Smith, F.~E.~V. Barrientos, P.~{\"O}gren, and D.~Kragic,
  ``An adaptive control approach for opening doors and drawers under
  uncertainties,'' {\em Transactions on Robotics (T-RO)}, vol.~32, no.~1,
  pp.~161--175, 2016.

\bibitem{wang2020learning}
J.~Wang, S.~Lin, C.~Hu, Y.~Zhu, and L.~Zhu, ``Learning semantic keypoint
  representations for door opening manipulation,'' {\em Robotics and Automation
  Letters (RA-L)}, vol.~5, no.~4, pp.~6980--6987, 2020.

\bibitem{shankar2016kinematics}
K.~Shankar, {\em Kinematics and Local Motion Planning for Quasi-static
  Whole-body Mobile Manipulation}.
\newblock PhD thesis, California Institute of Technology, 2016.

\bibitem{bodily2017motion}
D.~M. Bodily, T.~F. Allen, and M.~D. Killpack, ``Motion planning for mobile
  robots using inverse kinematics branching,'' in {\em Proceedings of
  International Conference on Robotics and Automation (ICRA)}, 2017.

\bibitem{chitta2010planning}
S.~Chitta, B.~Cohen, and M.~Likhachev, ``Planning for autonomous door opening
  with a mobile manipulator,'' in {\em Proceedings of International Conference
  on Robotics and Automation (ICRA)}, 2010.

\bibitem{gallagher2006body}
S.~Gallagher, {\em How the body shapes the mind}.
\newblock Clarendon Press, 2006.

\bibitem{holmes2006beyond}
N.~P. Holmes and C.~Spence, ``Beyond the body schema: Visual, prosthetic, and
  technological contributions to bodily perception and awareness,'' {\em Human
  body perception from the inside out: Advances in visual cognition},
  pp.~15--64, 2006.

\bibitem{hoffmann2010body}
M.~Hoffmann, H.~Marques, A.~Arieta, H.~Sumioka, M.~Lungarella, and R.~Pfeifer,
  ``Body schema in robotics: a review,'' {\em IEEE Transactions on Autonomous
  Mental Development}, vol.~2, no.~4, pp.~304--324, 2010.

\bibitem{likar2014virtual}
N.~Likar, B.~Nemec, and L.~{\v{Z}}lajpah, ``Virtual mechanism approach for
  dual-arm manipulation,'' {\em Robotica}, vol.~32, no.~6, 2014.

\bibitem{pratt1997virtual}
J.~Pratt, P.~Dilworth, and G.~Pratt, ``Virtual model control of a bipedal
  walking robot,'' in {\em Proceedings of International Conference on Robotics
  and Automation (ICRA)}, 1997.

\bibitem{pratt2001virtual}
J.~Pratt, C.-M. Chew, A.~Torres, P.~Dilworth, and G.~Pratt, ``Virtual model
  control: An intuitive approach for bipedal locomotion,'' {\em International
  Journal of Robotics Research (IJRR)}, vol.~20, no.~2, pp.~129--143, 2001.

\bibitem{wang2015cooperative}
Y.~Wang, C.~Smith, Y.~Karayiannidis, and P.~{\"O}gren, ``Cooperative control of
  a serial-to-parallel structure using a virtual kinematic chain in a mobile
  dual-arm manipulation application,'' in {\em Proceedings of International
  Conference on Intelligent Robots and Systems (IROS)}, 2015.

\bibitem{wang2016whole}
Y.~Wang, C.~Smith, Y.~Karayiannidis, and P.~{\"O}gren, ``Whole body control of
  a dual-arm mobile robot using a virtual kinematic chain,'' {\em International
  Journal of Humanoid Robotics}, vol.~13, no.~01, 2016.

\bibitem{laurenzi2019augmented}
A.~Laurenzi, E.~M. Hoffman, M.~P. Polverini, and N.~Tsagarakis, ``An augmented
  kinematic model for the cartesian control of the hybrid wheeled-legged
  quadrupedal robot centauro,'' {\em Robotics and Automation Letters (RA-L)},
  2019.

\bibitem{hart1968formal}
P.~E. Hart, N.~J. Nilsson, and B.~Raphael, ``A formal basis for the heuristic
  determination of minimum cost paths,'' {\em IEEE Transactions on Systems
  Science and Cybernetics}, vol.~4, no.~2, pp.~100--107, 1968.

\bibitem{stentz1997optimal}
A.~Stentz, ``Optimal and efficient path planning for partially known
  environments,'' in {\em Intelligent unmanned ground vehicles}, pp.~203--220,
  Springer, 1997.

\bibitem{lavalle2000rapidly}
S.~M. Lavalle and J.~J. Kuffner~Jr, ``Rapidly-exploring random trees: Progress
  and prospects,'' in {\em Intenational Workshop on Algorithmic and
  Computational Robotics}, 2000.

\bibitem{kuffner2000rrt}
J.~J. Kuffner and S.~M. LaValle, ``Rrt-connect: An efficient approach to
  single-query path planning,'' in {\em Proceedings of International Conference
  on Robotics and Automation (ICRA)}, 2000.

\bibitem{karaman2010incremental}
S.~Karaman and E.~Frazzoli, ``Incremental sampling-based algorithms for optimal
  motion planning,'' {\em Robotics Science and Systems VI}, vol.~104, no.~2,
  2010.

\bibitem{ratliff2009chomp}
N.~Ratliff, M.~Zucker, J.~A. Bagnell, and S.~Srinivasa, ``Chomp: gradient
  optimization techniques for efficient motion planning,'' in {\em Proceedings
  of International Conference on Robotics and Automation (ICRA)}, 2009.

\bibitem{schulman2014motion}
J.~Schulman, Y.~Duan, J.~Ho, A.~Lee, I.~Awwal, H.~Bradlow, J.~Pan, S.~Patil,
  K.~Goldberg, and P.~Abbeel, ``Motion planning with sequential convex
  optimization and convex collision checking,'' {\em International Journal of
  Robotics Research (IJRR)}, vol.~33, no.~9, pp.~1251--1270, 2014.

\bibitem{stuede2019door}
M.~Stuede, K.~Nuelle, S.~Tappe, and T.~Ortmaier, ``Door opening and traversal
  with an industrial cartesian impedance controlled mobile robot,'' in {\em
  Proceedings of International Conference on Robotics and Automation (ICRA)},
  2019.

\bibitem{gochev2012planning}
K.~Gochev, A.~Safonova, and M.~Likhachev, ``Planning with adaptive
  dimensionality for mobile manipulation,'' in {\em Proceedings of
  International Conference on Robotics and Automation (ICRA)}, 2012.

\bibitem{burget2013whole}
F.~Burget, A.~Hornung, and M.~Bennewitz, ``Whole-body motion planning for
  manipulation of articulated objects,'' in {\em Proceedings of International
  Conference on Robotics and Automation (ICRA)}, 2013.

\bibitem{toussaint2018differentiable}
M.~Toussaint, K.~Allen, K.~A. Smith, and J.~B. Tenenbaum, ``Differentiable
  physics and stable modes for tool-use and manipulation planning,'' in {\em
  Proceedings of Robotics: Science and Systems (RSS)}, 2018.

\bibitem{smits2011kdl}
R.~Smits, H.~Bruyninckx, and E.~Aertbeli{\"e}n, ``Kdl: Kinematics and dynamics
  library,'' 2011.
\newblock \url{https://www.orocos.org/kdl}.

\bibitem{rosmann2017kinodynamic}
C.~R{\"o}smann, F.~Hoffmann, and T.~Bertram, ``Kinodynamic trajectory
  optimization and control for car-like robots,'' in {\em Proceedings of
  International Conference on Intelligent Robots and Systems (IROS)}, 2017.

\bibitem{armstrong_2018}
L.~Armstrong, ``Optimization motion planning with tesseract and trajopt for
  industrial applications - ros-industrial.''

\bibitem{magyar2019timed}
B.~Magyar, N.~Tsiogkas, J.~Deray, S.~Pfeiffer, and D.~Lane, ``Timed-elastic
  bands for manipulation motion planning,'' {\em Robotics and Automation
  Letters (RA-L)}, vol.~4, no.~4, pp.~3513--3520, 2019.

\bibitem{jiao2021efficient}
Z.~Jiao, Z.~Zhang, W.~Wang, D.~Han, S.-C. Zhu, Y.~Zhu, and H.~Liu, ``Efficient
  task planning for mobile manipulation: a virtual kinematic chain
  perspective,'' in {\em Proceedings of International Conference on Intelligent
  Robots and Systems (IROS)}, 2021.

\bibitem{barragan2014interactive}
P.~R. Barrag{\"a}n, L.~P. Kaelbling, and T.~Lozano-P{\'e}rez, ``Interactive
  bayesian identification of kinematic mechanisms,'' in {\em Proceedings of
  International Conference on Robotics and Automation (ICRA)}, 2014.

\bibitem{paolillo2018interlinked}
A.~Paolillo, K.~Chappellet, A.~Bolotnikova, and A.~Kheddar, ``Interlinked
  visual tracking and robotic manipulation of articulated objects,'' {\em
  Robotics and Automation Letters (RA-L)}, vol.~3, no.~4, pp.~2746--2753, 2018.

\bibitem{martin2019coupled}
R.~Mart{\'\i}n-Mart{\'\i}n and O.~Brock, ``Coupled recursive estimation for
  online interactive perception of articulated objects,'' {\em International
  Journal of Robotics Research (IJRR)}, pp.~1--37, 2019.

\bibitem{abbatematteo2019learning}
B.~Abbatematteo, S.~Tellex, and G.~Konidaris, ``Learning to generalize
  kinematic models to novel objects,'' in {\em Conference on Robot Learning
  (CoRL)}, 2019.

\bibitem{han2021reconstructing}
M.~Han, Z.~Zhang, Z.~Jiao, X.~Xie, Y.~Zhu, S.-C. Zhu, and H.~Liu,
  ``Reconstructing interactive 3d scenes by panoptic mapping and cad model
  alignments,'' in {\em Proceedings of International Conference on Robotics and
  Automation (ICRA)}, 2021.

\bibitem{liang2019pointnetgpd}
H.~Liang, X.~Ma, S.~Li, M.~G{\"o}rner, S.~Tang, B.~Fang, F.~Sun, and J.~Zhang,
  ``Pointnetgpd: Detecting grasp configurations from point sets,'' in {\em
  Proceedings of International Conference on Robotics and Automation (ICRA)},
  2019.

\bibitem{mousavian20196}
A.~Mousavian, C.~Eppner, and D.~Fox, ``6-dof graspnet: Variational grasp
  generation for object manipulation,'' in {\em Proceedings of the IEEE
  Conference on Computer Vision and Pattern Recognition (CVPR)}, 2019.

\bibitem{ichnowski2020deep}
J.~Ichnowski, Y.~Avigal, V.~Satish, and K.~Goldberg, ``Deep learning can
  accelerate grasp-optimized motion planning,'' {\em Science Robotics}, vol.~5,
  no.~48, 2020.

\end{thebibliography}
}

\end{document}